# VIME: Variational Information Maximizing Exploration

**Rein Houthooft**[§†‡], **Xi Chen**[†‡], **Yan Duan**[†‡], **John Schulman**[†‡], **Filip De Turck**[§], **Pieter Abbeel**[†‡]

[†] UC Berkeley, Department of Electrical Engineering and Computer Sciences
[§] Ghent University - imec, Department of Information Technology
[‡] OpenAI

## Abstract

Scalable and effective exploration remains a key challenge in reinforcement learning (RL). While there are methods with optimality guarantees in the setting of discrete state and action spaces, these methods cannot be applied in high-dimensional deep RL scenarios. As such, most contemporary RL relies on simple heuristics such as $\epsilon$-greedy exploration or adding Gaussian noise to the controls. This paper introduces Variational Information Maximizing Exploration (VIME), an exploration strategy based on maximization of information gain about the agent's belief of environment dynamics. We propose a practical implementation, using variational inference in Bayesian neural networks which efficiently handles continuous state and action spaces. VIME modifies the MDP reward function, and can be applied with several different underlying RL algorithms. We demonstrate that VIME achieves significantly better performance compared to heuristic exploration methods across a variety of continuous control tasks and algorithms, including tasks with very sparse rewards.

## 1 Introduction

Reinforcement learning (RL) studies how an agent can maximize its cumulative reward in a previously unknown environment, which it learns about through experience. A long-standing problem is how to manage the trade-off between exploration and exploitation. In *exploration*, the agent experiments with novel strategies that may improve returns in the long run; in *exploitation*, it maximizes rewards through behavior that is known to be successful. An effective exploration strategy allows the agent to generate trajectories that are maximally informative about the environment. For small tasks, this trade-off can be handled effectively through Bayesian RL [1] and PAC-MDP methods [2–6], which offer formal guarantees. However, these guarantees assume discrete state and action spaces. Hence, in settings where state-action discretization is infeasible, many RL algorithms use heuristic exploration strategies. Examples include acting randomly using $\epsilon$-greedy or Boltzmann exploration [7], and utilizing Gaussian noise on the controls in policy gradient methods [8]. These heuristics often rely on random walk behavior which can be highly inefficient, for example Boltzmann exploration requires a training time exponential in the number of states in order to solve the well-known $n$-chain MDP [9]. In between formal methods and simple heuristics, several works have proposed to address the exploration problem using less formal, but more expressive methods [10–14]. However, none of them fully address exploration in continuous control, as discretization of the state-action space scales exponentially in its dimensionality. For example, the Walker2D task [15] has a 26-dim state-action space. If we assume a coarse discretization into 10 bins for each dimension, a table of state-action visitation counts would require $10^{26}$ entries.

This paper proposes a curiosity-driven exploration strategy, making use of information gain about the agent's internal belief of the dynamics model as a driving force. This principle can be traced back to the concepts of *curiosity* and *surprise* [16–18]. Within this framework, agents are encouraged to take actions that result in states they deem surprising—i.e., states that cause large updates to the dynamics model distribution. We propose a practical implementation of measuring information gain using variational inference. Herein, the agent's current understanding of the environment dynamics is represented by a Bayesian neural network (BNN) [19, 20]. We also show how this can be interpreted as measuring compression improvement, a proposed model of curiosity [21]. In contrast to previous curiosity-based approaches [10, 22], our model scales naturally to continuous state and action spaces. The presented approach is evaluated on a range of continuous control tasks, and multiple underlying RL algorithms. Experimental results show that VIME achieves significantly better performance than naïve exploration strategies.

## 2 Methodology

In Section 2.1, we establish notation for the subsequent equations. Next, in Section 2.2, we explain the theoretical foundation of curiosity-driven exploration. In Section 2.3 we describe how to adapt this idea to continuous control, and we show how to build on recent advances in variational inference for Bayesian neural networks (BNNs) to make this formulation practical. Thereafter, we make explicit the intuitive link between compression improvement and the variational lower bound in Section 2.4. Finally, Section 2.5 describes how our method is practically implemented.

### 2.1 Preliminaries

This paper assumes a finite-horizon discounted Markov decision process (MDP), defined by $(\mathcal{S}, \mathcal{A}, \mathcal{P}, r, \rho_0, \gamma, T)$, in which $\mathcal{S} \subseteq \mathbb{R}^n$ is a state set, $\mathcal{A} \subseteq \mathbb{R}^m$ an action set, $\mathcal{P} : \mathcal{S} \times \mathcal{A} \times \mathcal{S} \to \mathbb{R}_{\geq 0}$ a transition probability distribution, $r : \mathcal{S} \times \mathcal{A} \to \mathbb{R}$ a bounded reward function, $\rho_0 : \mathcal{S} \to \mathbb{R}_{\geq 0}$ an initial state distribution, $\gamma \in (0, 1]$ a discount factor, and $T$ the horizon. States and actions viewed as random variables are abbreviated as $S$ and $A$. The presented models are based on the optimization of a stochastic policy $\pi_\alpha : \mathcal{S} \times \mathcal{A} \to \mathbb{R}_{\geq 0}$, parametrized by $\alpha$. Let $\mu(\pi_\alpha)$ denote its expected discounted return: $\mu(\pi_\alpha) = \mathbb{E}_\tau[\sum_{t=0}^T \gamma^t r(s_t, a_t)]$, where $\tau = (s_0, a_0, \ldots)$ denotes the whole trajectory, $s_0 \sim \rho_0(s_0)$, $a_t \sim \pi_\alpha(a_t|s_t)$, and $s_{t+1} \sim \mathcal{P}(s_{t+1}|s_t, a_t)$.

### 2.2 Curiosity

Our method builds on the theory of curiosity-driven exploration [16, 17, 21, 22], in which the agent engages in systematic exploration by seeking out state-action regions that are relatively unexplored. The agent models the environment dynamics via a model $p(s_{t+1}|s_t, a_t; \theta)$, parametrized by the random variable $\Theta$ with values $\theta \in \Theta$. Assuming a prior $p(\theta)$, it maintains a distribution over dynamic models through a distribution over $\theta$, which is updated in a Bayesian manner (as opposed to a point estimate). The history of the agent up until time step $t$ is denoted as $\xi_t = \{s_1, a_1, \ldots, s_t\}$. According to curiosity-driven exploration [17], the agent should take actions that maximize the reduction in uncertainty about the dynamics. This can be formalized as maximizing the sum of reductions in entropy

$$\sum_t \left( H(\Theta|\xi_t, a_t) - H(\Theta|S_{t+1}, \xi_t, a_t) \right), \tag{1}$$

through a sequence of actions $\{a_t\}$. According to information theory, the individual terms equal the mutual information between the next state distribution $S_{t+1}$ and the model parameter $\Theta$, namely $I(S_{t+1}; \Theta|\xi_t, a_t)$. Therefore, the agent is encouraged to take actions that lead to states that are maximally informative about the dynamics model. Furthermore, we note that

$$I(S_{t+1}; \Theta|\xi_t, a_t) = \mathbb{E}_{s_{t+1} \sim \mathcal{P}(\cdot|\xi_t, a_t)} \left[ D_{\mathrm{KL}}[p(\theta|\xi_t, a_t, s_{t+1}) \,\|\, p(\theta|\xi_t)] \right], \tag{2}$$

the KL divergence from the agent's new belief over the dynamics model to the old one, taking expectation over all possible next states according to the true dynamics $\mathcal{P}$. This KL divergence can be interpreted as *information gain*.

If calculating the posterior dynamics distribution is tractable, it is possible to optimize Eq. (2) directly by maintaining a belief over the dynamics model [17]. However, this is not generally the case. Therefore, a common practice [10, 23] is to use RL to approximate planning for maximal mutual information along a trajectory $\sum_t I(S_{t+1}; \Theta | \xi_t, a_t)$ by adding each term $I(S_{t+1}; \Theta | \xi_t, a_t)$ as an *intrinsic reward*, which captures the agent's surprise in the form of a reward function. This is practically realized by taking actions $a_t \sim \pi_\alpha(s_t)$ and sampling $s_{t+1} \sim \mathcal{P}(\cdot | s_t, a_t)$ in order to add $D_{\text{KL}}[p(\theta|\xi_t, a_t, s_{t+1}) \| p(\theta|\xi_t)]$ to the external reward. The trade-off between exploitation and exploration can now be realized explicitly as follows:

$$r'(s_t, a_t, s_{t+1}) = r(s_t, a_t) + \eta D_{\text{KL}}[p(\theta|\xi_t, a_t, s_{t+1}) \| p(\theta|\xi_t)], \tag{3}$$

with $\eta \in \mathbb{R}_+$ a hyperparameter controlling the urge to explore. In conclusion, the biggest practical issue with maximizing information gain for exploration is that the computation of Eq. (3) requires calculating the posterior $p(\theta|\xi_t, a_t, s_{t+1})$, which is generally intractable.

### 2.3 Variational Bayes

We propose a tractable solution to maximize the information gain objective presented in the previous section. In a purely Bayesian setting, we can derive the posterior distribution given a new state-action pair through Bayes' rule as

$$p(\theta|\xi_t, a_t, s_{t+1}) = \frac{p(\theta|\xi_t) p(s_{t+1}|\xi_t, a_t; \theta)}{p(s_{t+1}|\xi_t, a_t)}, \tag{4}$$

with $p(\theta|\xi_t, a_t) = p(\theta|\xi_t)$ as actions do not influence beliefs about the environment [17]. Herein, the denominator is computed through the integral

$$p(s_{t+1}|\xi_t, a_t) = \int_\Theta p(s_{t+1}|\xi_t, a_t; \theta) p(\theta|\xi_t) d\theta. \tag{5}$$

In general, this integral tends to be intractable when using highly expressive parametrized models (e.g., neural networks), which are often needed to accurately capture the environment model in high-dimensional continuous control.

We propose a practical solution through variational inference [24]. Herein, we embrace the fact that calculating the posterior $p(\theta|\mathcal{D})$ for a data set $\mathcal{D}$ is intractable. Instead we approximate it through an alternative distribution $q(\theta; \phi)$, parametrized by $\phi$, by minimizing $D_{\text{KL}}[q(\theta; \phi) \| p(\theta|\mathcal{D})]$. This is done through maximization of the *variational lower bound* $L[q(\theta; \phi), \mathcal{D}]$:

$$L[q(\theta; \phi), \mathcal{D}] = \mathbb{E}_{\theta \sim q(\cdot; \phi)}[\log p(\mathcal{D}|\theta)] - D_{\text{KL}}[q(\theta; \phi) \| p(\theta)]. \tag{6}$$

Rather than computing information gain in Eq. (3) explicitly, we compute an approximation to it, leading to the following total reward:

$$r'(s_t, a_t, s_{t+1}) = r(s_t, a_t) + \eta D_{\text{KL}}[q(\theta; \phi_{t+1}) \| q(\theta; \phi_t)], \tag{7}$$

with $\phi_{t+1}$ the updated and $\phi_t$ the old parameters representing the agent's belief. Natural candidates for parametrizing the agent's dynamics model are Bayesian neural networks (BNNs) [19], as they maintain a distribution over their weights. This allows us to view the BNN as an infinite neural network ensemble by integrating out its parameters:

$$p(y|x) = \int_\Theta p(y|x; \theta) q(\theta; \phi) d\theta. \tag{8}$$

In particular, we utilize a BNN parametrized by a fully factorized Gaussian distribution [20]. Practical BNN implementation details are deferred to Section 2.5, while we give some intuition into the behavior of BNNs in the appendix.

### 2.4 Compression

It is possible to derive an interesting relationship between compression improvement—an intrinsic reward objective defined in [25], and the information gain of Eq. (2). In [25], the agent's curiosity is equated with compression improvement, measured through $C(\xi_t; \phi_{t-1}) - C(\xi_t; \phi_t)$, where $C(\xi; \phi)$ is the description length of $\xi$ using $\phi$ as a model. Furthermore, it is known that the negative variational lower bound can be viewed as the description length [19]. Hence, we can write compression improvement as $L[q(\theta; \phi_t), \xi_t] - L[q(\theta; \phi_{t-1}), \xi_t]$. In addition, an alternative formulation of the variational lower bound in Eq. (6) is given by

$$\log p(\mathcal{D}) = \overbrace{\int_{\Theta} q(\theta; \phi) \log \frac{p(\theta, \mathcal{D})}{q(\theta; \phi)} d\theta}^{L[q(\theta; \phi), \mathcal{D}]} + D_{\mathrm{KL}}[q(\theta; \phi) \,\|\, p(\theta | \mathcal{D})]. \tag{9}$$

Thus, compression improvement can now be written as

$$\left(\log p(\xi_t) - D_{\mathrm{KL}}[q(\theta; \phi_t) \,\|\, p(\theta | \xi_t)]\right) - \left(\log p(\xi_t) - D_{\mathrm{KL}}[q(\theta; \phi_{t-1}) \,\|\, p(\theta | \xi_t)]\right). \tag{10}$$

If we assume that $\phi_t$ perfectly optimizes the variational lower bound for the history $\xi_t$, then $D_{\mathrm{KL}}[q(\theta; \phi_t) \,\|\, p(\theta | \xi_t)] = 0$, which occurs when the approximation equals the true posterior, i.e., $q(\theta; \phi_t) = p(\theta | \xi_t)$. Hence, compression improvement becomes $D_{\mathrm{KL}}[p(\theta | \xi_{t-1}) \,\|\, p(\theta | \xi_t)]$. Therefore, optimizing for compression improvement comes down to optimizing the KL divergence from the posterior given the past history $\xi_{t-1}$ to the posterior given the total history $\xi_t$. As such, we arrive at an alternative way to encode curiosity than information gain, namely $D_{\mathrm{KL}}[p(\theta | \xi_t) \,\|\, p(\theta | \xi_t, a_t, s_{t+1})]$, its reversed KL divergence. In experiments, we noticed no significant difference between the two KL divergence variants. This can be explained as both variants are locally equal when introducing small changes to the parameter distributions. Investigation of how to combine both information gain and compression improvement is deferred to future work.

### 2.5 Implementation

The complete method is summarized in Algorithm 1. We first set forth implementation and parametrization details of the dynamics BNN. The BNN weight distribution $q(\theta; \phi)$ is given by the fully factorized Gaussian distribution [20]:

$$q(\theta; \phi) = \prod_{i=1}^{|\Theta|} \mathcal{N}(\theta_i | \mu_i; \sigma_i^2). \tag{11}$$

Hence, $\phi = \{\mu, \sigma\}$, with $\mu$ the Gaussian's mean vector and $\sigma$ the covariance matrix diagonal. This is particularly convenient as it allows for a simple analytical formulation of the KL divergence. This is described later in this section. Because of the restriction $\sigma > 0$, the standard deviation of the Gaussian BNN parameter is parametrized as $\sigma = \log(1 + e^\rho)$, with $\rho \in \mathbb{R}$ [20].

Now the training of the dynamics BNN through optimization of the variational lower bound is described. The second term in Eq. (6) is approximated through sampling $\mathbb{E}_{\theta \sim q(\cdot; \phi)}[\log p(\mathcal{D} | \theta)] \approx \frac{1}{N} \sum_{i=1}^{N} \log p(\mathcal{D} | \theta_i)$ with $N$ samples drawn according to $\theta \sim q(\cdot; \phi)$ [20]. Optimizing the variational lower bound in Eq. (6) in combination with the reparametrization trick is called stochastic gradient variational Bayes (SGVB) [26] or Bayes by Backprop [20]. Furthermore, we make use of the local reparametrization trick proposed in [26], in which sampling at the weights is replaced by sampling the neuron pre-activations, which is more computationally efficient and reduces gradient variance. The optimization of the variational lower bound is done at regular intervals during the RL training process, by sampling $\mathcal{D}$ from a FIFO replay pool that stores recent samples $(s_t, a_t, s_{t+1})$. This is to break up the strong intratrajectory sample correlation which destabilizes learning in favor of obtaining i.i.d. data [7]. Moreover, it diminishes the effect of compounding posterior approximation errors.

The posterior distribution of the dynamics parameter, which is needed to compute the KL divergence in the total reward function $r'$ of Eq. (7), can be computed through the following minimization

$$\phi' = \arg\min_{\phi} \left[ \underbrace{\underbrace{D_{\mathrm{KL}}[q(\theta; \phi) \,\|\, q(\theta; \phi_{t-1})]}_{\ell_{\mathrm{KL}}(q(\theta; \phi))} - \mathbb{E}_{\theta \sim q(\cdot; \phi)}[\log p(s_t | \xi_t, a_t; \theta)]}_{\ell(q(\theta; \phi), s_t)} \right], \tag{12}$$

where we replace the expectation over $\theta$ with samples $\theta \sim q(\cdot; \phi)$. Because we only update the model periodically based on samples drawn from the replay pool, this optimization can be performed in parallel for each $s_t$, keeping $\phi_{t-1}$ fixed. Once $\phi'$ has been obtained, we can use it to compute the intrinsic reward.

**Algorithm 1:** Variational Information Maximizing Exploration (VIME)

**for** each epoch $n$ **do**
  **for** each timestep $t$ in each trajectory generated during $n$ **do**
    Generate action $a_t \sim \pi_\alpha(s_t)$ and sample state $s_{t+1} \sim \mathcal{P}(\cdot|\xi_t, a_t)$, get $r(s_t, a_t)$.
    Add triplet $(s_t, a_t, s_{t+1})$ to FIFO replay pool $\mathcal{R}$.
    Compute $D_{\mathrm{KL}}[q(\theta; \phi'_{n+1}) \| q(\theta; \phi_{n+1})]$ by approximation $\nabla^\top H^{-1} \nabla$, following Eq. (16) for diagonal BNNs, or by optimizing Eq. (12) to obtain $\phi'_{n+1}$ for general BNNs.
    Divide $D_{\mathrm{KL}}[q(\theta; \phi'_{n+1}) \| q(\theta; \phi_{n+1})]$ by median of previous KL divergences.
    Construct $r'(s_t, a_t, s_{t+1}) \leftarrow r(s_t, a_t) + \eta D_{\mathrm{KL}}[q(\theta; \phi'_{n+1}) \| q(\theta; \phi_{n+1})]$, following Eq. (7).
  Minimize $D_{\mathrm{KL}}[q(\theta; \phi_n) \| p(\theta)] - \mathbb{E}_{\theta \sim q(\cdot; \phi_n)}[\log p(\mathcal{D}|\theta)]$ following Eq. (6), with $\mathcal{D}$ sampled randomly from $\mathcal{R}$, leading to updated posterior $q(\theta; \phi_{n+1})$.
  Use rewards $\{r'(s_t, a_t, s_{t+1})\}$ to update policy $\pi_\alpha$ using any standard RL method.

To optimize Eq. (12) efficiently, we only take a single second-order step. This way, the gradient is rescaled according to the curvature of the KL divergence at the origin. As such, we compute $D_{\mathrm{KL}}[q(\theta; \phi + \lambda \Delta \phi) \| q(\theta; \phi)]$, with the update step $\Delta \phi$ defined as

$$\Delta \phi = H^{-1}(\ell) \nabla_\phi \ell(q(\theta; \phi), s_t), \tag{13}$$

in which $H(\ell)$ is the Hessian of $\ell(q(\theta; \phi), s_t)$. Since we assume that the variational approximation is a fully factorized Gaussian, the KL divergence from posterior to prior has a particularly simple form:

$$D_{\mathrm{KL}}[q(\theta; \phi) \| q(\theta; \phi')] = \tfrac{1}{2} \sum_{i=1}^{|\Theta|} \left( \left( \tfrac{\sigma_i}{\sigma'_i} \right)^2 + 2\log \sigma'_i - 2\log \sigma_i + \tfrac{(\mu'_i - \mu_i)^2}{\sigma'^2_i} \right) - \tfrac{|\Theta|}{2}. \tag{14}$$

Because this KL divergence is approximately quadratic in its parameters and the log-likelihood term can be seen as locally linear compared to this highly curved KL term, we approximate $H$ by only calculating it for the term KL term $\ell_{\mathrm{KL}}(q(\theta; \phi))$. This can be computed very efficiently in case of a fully factorized Gaussian distribution, as this approximation becomes a diagonal matrix. Looking at Eq. (14), we can calculate the following Hessian at the origin. The $\mu$ and $\rho$ entries are defined as

$$\frac{\partial^2 \ell_{\mathrm{KL}}}{\partial \mu_i^2} = \frac{1}{\log^2(1 + e^{\rho_i})} \quad \text{and} \quad \frac{\partial^2 \ell_{\mathrm{KL}}}{\partial \rho_i^2} = \frac{2e^{2\rho_i}}{(1 + e^{\rho_i})^2} \frac{1}{\log^2(1 + e^{\rho_i})}, \tag{15}$$

while all other entries are zero. Furthermore, it is also possible to approximate the KL divergence through a second-order Taylor expansion as $\tfrac{1}{2} \Delta \phi H \Delta \phi = \tfrac{1}{2} \left( H^{-1} \nabla \right)^\top H \left( H^{-1} \nabla \right)$, since both the value and gradient of the KL divergence are zero at the origin. This gives us

$$D_{\mathrm{KL}}[q(\theta; \phi + \lambda \Delta \phi) \| q(\theta; \phi)] \approx \tfrac{1}{2} \lambda^2 \nabla_\phi \ell^\top H^{-1}(\ell_{\mathrm{KL}}) \nabla_\phi \ell. \tag{16}$$

Note that $H^{-1}(\ell_{\mathrm{KL}})$ is diagonal, so this expression can be computed efficiently. Instead of using the KL divergence $D_{\mathrm{KL}}[q(\theta; \phi_{t+1}) \| q(\theta; \phi_t)]$ directly as an intrinsic reward in Eq. (7), we normalize it by division through the average of the median KL divergences taken over a fixed number of previous trajectories. Rather than focusing on its absolute value, we emphasize relative difference in KL divergence between samples. This accomplishes the same effect since the variance of KL divergence converges to zero, once the model is fully learned.

## 3 Experiments

In this section, we investigate (i) whether VIME can succeed in domains that have extremely sparse rewards, (ii) whether VIME improves learning when the reward is shaped to guide the agent towards its goal, and (iii) how $\eta$, as used in in Eq. (3), trades off exploration and exploitation behavior. All experiments make use of the rllab [15] benchmark code base and the complementary continuous control tasks suite. The following tasks are part of the experimental setup: CartPole ($\mathcal{S} \subseteq \mathbb{R}^4$, $\mathcal{A} \subseteq \mathbb{R}^1$), CartPoleSwingup ($\mathcal{S} \subseteq \mathbb{R}^4$, $\mathcal{A} \subseteq \mathbb{R}^1$), DoublePendulum ($\mathcal{S} \subseteq \mathbb{R}^6$, $\mathcal{A} \subseteq \mathbb{R}^1$), MountainCar ($\mathcal{S} \subseteq \mathbb{R}^3$, $\mathcal{A} \subseteq \mathbb{R}^1$), locomotion tasks HalfCheetah ($\mathcal{S} \subseteq \mathbb{R}^{20}$, $\mathcal{A} \subseteq \mathbb{R}^6$), Walker2D ($\mathcal{S} \subseteq \mathbb{R}^{20}$, $\mathcal{A} \subseteq \mathbb{R}^6$), and the hierarchical task SwimmerGather ($\mathcal{S} \subseteq \mathbb{R}^{33}$, $\mathcal{A} \subseteq \mathbb{R}^2$).

Performance is measured through the average return (not including the intrinsic rewards) over the trajectories generated (y-axis) at each iteration (x-axis). More specifically, the darker-colored lines in each plot represent the median performance over a fixed set of 10 random seeds while the shaded areas show the interquartile range at each iteration. Moreover, the number in each legend shows this performance measure, averaged over all iterations. The exact setup is described in the Appendix.

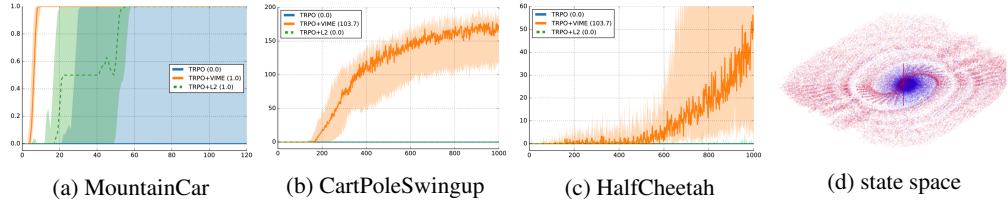

(a) MountainCar  (b) CartPoleSwingup  (c) HalfCheetah  (d) state space

Figure 1: (a,b,c) TRPO+VIME versus TRPO on tasks with sparse rewards; (d) comparison of TRPO+VIME (red) and TRPO (blue) on MountainCar: visited states until convergence

Domains with sparse rewards are difficult to solve through naïve exploration behavior because, before the agent obtains any reward, it lacks a feedback signal on how to improve its policy. This allows us to test whether an exploration strategy is truly capable of systematic exploration, rather than improving existing RL algorithms by adding more hyperparameters. Therefore, VIME is compared with heuristic exploration strategies on the following tasks with sparse rewards. A reward of $+1$ is given when the car escapes the valley on the right side in MountainCar; when the pole is pointed upwards in CartPoleSwingup; and when the cheetah moves forward over five units in HalfCheetah. We compare VIME with the following baselines: only using Gaussian control noise [15] and using the $\ell^2$ BNN prediction error as an intrinsic reward, a continuous extension of [10]. TRPO [8] is used as the RL algorithm, as it performs very well compared to other methods [15]. Figure 1 shows the performance results. We notice that using a naïve exploration performs very poorly, as it is almost never able to reach the goal in any of the tasks. Similarly, using $\ell^2$ errors does not perform well. In contrast, VIME performs much better, achieving the goal in most cases. This experiment demonstrates that curiosity drives the agent to explore, even in the absence of *any* initial reward, where naïve exploration completely breaks down.

To further strengthen this point, we have evaluated VIME on the highly difficult hierarchical task SwimmerGather in Figure 5 whose reward signal is naturally sparse. In this task, a two-link robot needs to reach "apples" while avoiding "bombs" that are perceived through a laser scanner. However, before it can make any forward progress, it has to learn complex locomotion primitives in the absence of any reward. None of the RL methods tested previously in [15] were able to make progress with naïve exploration. Remarkably, VIME leads the agent to acquire coherent motion primitives without any reward guidance, achieving promising results on this challenging task.

Next, we investigate whether VIME is widely applicable by (i) testing it on environments where the reward is well shaped, and (ii) pairing it with different RL methods. In addition to TRPO, we choose to equip REINFORCE [27] and ERWR [28] with VIME because these two algorithms usually suffer from premature convergence to suboptimal policies [15, 29], which can potentially be alleviated by better exploration. Their performance is shown in Figure 2 on several well-established continuous control tasks. Furthermore, Figure 3 shows the same comparison for the Walker2D locomotion task. In the majority of cases, VIME leads to a significant performance gain over heuristic exploration. Our exploration method allows the RL algorithms to converge faster, and notably helps REINFORCE and ERWR avoid converging to a locally optimal solution on DoublePendulum and MountainCar. We note that in environments such as CartPole, a better exploration strategy is redundant as following the policy gradient direction leads to the globally optimal solution. Additionally, we tested adding Gaussian noise to the rewards as a baseline, which did not improve performance.

To give an intuitive understanding of VIME's exploration behavior, the distribution of visited states for both naïve exploration and VIME after convergence is investigated. Figure 1d shows that using Gaussian control noise exhibits random walk behavior: the state visitation plot is more condensed and ball-shaped around the center. In comparison, VIME leads to a more diffused visitation pattern, exploring the states more efficiently, and hence reaching the goal more quickly.

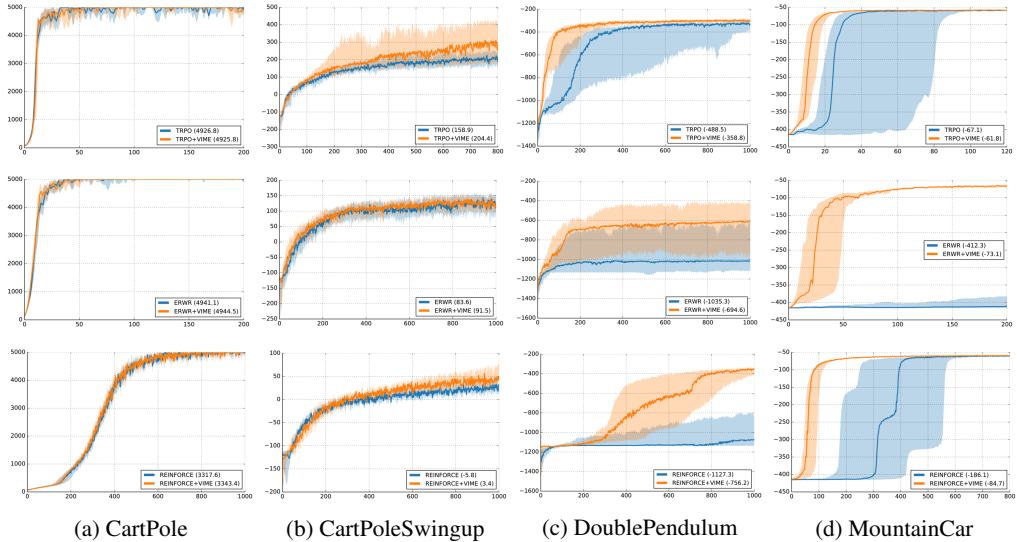

(a) CartPole  (b) CartPoleSwingup  (c) DoublePendulum  (d) MountainCar

Figure 2: Performance of TRPO (top row), ERWR (middle row), and REINFORCE (bottom row) with (+VIME) and without exploration for different continuous control tasks.

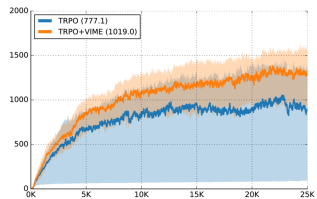
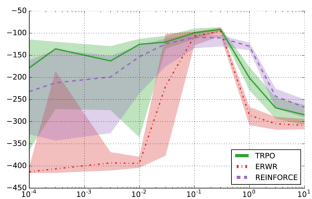
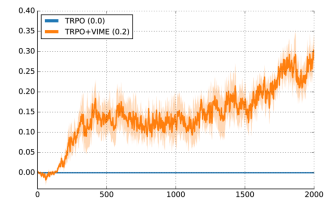

Figure 3: Performance of TRPO with and without VIME on the high-dimensional Walker2D locomotion task.

Figure 4: VIME: performance over the first few iterations for TRPO, REINFORCE, and ERWR i.f.o. $\eta$ on MountainCar.

Figure 5: Performance of TRPO with and without VIME on the challenging hierarchical task SwimmerGather.

Finally, we investigate how $\eta$, as used in in Eq. (3), trades off exploration and exploitation behavior. On the one hand, higher $\eta$ values should lead to a higher curiosity drive, causing more exploration. On the other hand, very low $\eta$ values should reduce VIME to traditional Gaussian control noise. Figure 4 shows the performance on MountainCar for different $\eta$ values. Setting $\eta$ too high clearly results in prioritizing exploration over getting additional external reward. Too low of an $\eta$ value reduces the method to the baseline algorithm, as the intrinsic reward contribution to the total reward $r'$ becomes negligible. Most importantly, this figure highlights that there is a wide $\eta$ range for which the task is best solved, across different algorithms.

## 4 Related Work

A body of theoretically oriented work demonstrates exploration strategies that are able to learn online in a previously unknown MDP and incur a polynomial amount of regret—as a result, these algorithms find a near-optimal policy in a polynomial amount of time. Some of these algorithms are based on the principle of optimism under uncertainty: $E^3$ [3], R-Max [4], UCRL [30]. An alternative approach is Bayesian reinforcement learning methods, which maintain a distribution over possible MDPs [1, 17, 23, 31]. The optimism-based exploration strategies have been extended to continuous state spaces, for example, [6, 9], however these methods do not accommodate nonlinear function approximators.

Practical RL algorithms often rely on simple exploration heuristics, such as $\epsilon$-greedy and Boltzmann exploration [32]. However, these heuristics exhibit random walk exploratory behavior, which can lead to exponential regret even in case of small MDPs [9]. Our proposed method of utilizing information gain can be traced back to [22], and has been further explored in [17, 33, 34]. Other metrics for curiosity have also been proposed, including prediction error [10, 35], prediction error improvement [36], leverage [14], neuro-correlates [37], and predictive information [38]. These methods have not been applied directly to high-dimensional continuous control tasks without discretization. We refer the reader to [21, 39] for an extensive review on curiosity and intrinsic rewards.

Recently, there have been various exploration strategiesproposed in the context of deep RL. [10] proposes to use the $\ell^2$ prediction error as the intrinsic reward. [12] performs approximate visitation counting in a learned state embedding using Gaussian kernels. [11] proposes a form of Thompson sampling, training multiple value functions using bootstrapping. Although these approaches can scale up to high-dimensional state spaces, they generally assume discrete action spaces. [40] make use of mutual information for gait stabilization in continuous control, but rely on state discretization. Finally, [41] proposes a variational method for information maximization in the context of optimizing *empowerment*, which, as noted by [42], does not explicitly favor exploration.

## 5 Conclusions

We have proposed Variational Information Maximizing Exploration (VIME), a curiosity-driven exploration strategy for continuous control tasks. Variational inference is used to approximate the posterior distribution of a Bayesian neural network that represents the environment dynamics. Using information gain in this learned dynamics model as intrinsic rewards allows the agent to optimize for both external reward and intrinsic surprise simultaneously. Empirical results show that VIME performs significantly better than heuristic exploration methods across various continuous control tasks and algorithms. As future work, we would like to investigate measuring surprise in the value function and using the learned dynamics model for planning.

## Acknowledgments

This work was done in collaboration between UC Berkeley, Ghent University and OpenAI. The work at Berkeley was supported in part by the DARPA SIMPLEX program and by ONR through a PECASE award. Xi Chen was supported by a Berkeley AI Research lab Fellowship. Yan Duan was supported by a Berkeley AI Research lab Fellowship and a Berkeley AI Huawei Fellowship. Rein Houthooft was supported by a Ph.D. Fellowship of the Research Foundation - Flanders (FWO).## References

[1] M. Ghavamzadeh, S. Mannor, J. Pineau, and A. Tamar, "Bayesian reinforcement learning: A survey", *Found. Trends. Mach. Learn.*, vol. 8, no. 5-6, pp. 359–483, 2015.

[2] S. Kakade, M. Kearns, and J. Langford, "Exploration in metric state spaces", in *ICML*, vol. 3, 2003, pp. 306–312.

[3] M. Kearns and S. Singh, "Near-optimal reinforcement learning in polynomial time", *Mach. Learn.*, vol. 49, no. 2-3, pp. 209–232, 2002.

[4] R. I. Brafman and M. Tennenholtz, "R-Max - a general polynomial time algorithm for near-optimal reinforcement learning", *J. Mach. Learn. Res.*, vol. 3, pp. 213–231, 2003.

[5] P. Auer, "Using confidence bounds for exploitation-exploration trade-offs", *J. Mach. Learn. Res.*, vol. 3, pp. 397–422, 2003.

[6] J. Pazis and R. Parr, "PAC optimal exploration in continuous space Markov decision processes", in *AAAI*, 2013.

[7] V. Mnih, K. Kavukcuoglu, D. Silver, A. A. Rusu, J. Veness, M. G. Bellemare, A. Graves, M. Riedmiller, A. K. Fidjeland, G. Ostrovski, *et al.*, "Human-level control through deep reinforcement learning", *Nature*, vol. 518, no. 7540, pp. 529–533, 2015.

[8] J. Schulman, S. Levine, P. Moritz, M. I. Jordan, and P. Abbeel, "Trust region policy optimization", in *ICML*, 2015.

[9] I. Osband, B. Van Roy, and Z. Wen, "Generalization and exploration via randomized value functions", *ArXiv preprint arXiv:1402.0635*, 2014.

[10] B. C. Stadie, S. Levine, and P. Abbeel, "Incentivizing exploration in reinforcement learning with deep predictive models", *ArXiv preprint arXiv:1507.00814*, 2015.
[11] I. Osband, C. Blundell, A. Pritzel, and B. Van Roy, "Deep exploration via bootstrapped DQN", in *NIPS*, 2016.
[12] J. Oh, X. Guo, H. Lee, R. L. Lewis, and S. Singh, "Action-conditional video prediction using deep networks in Atari games", in *NIPS*, 2015, pp. 2845–2853.
[13] T. Hester and P. Stone, "Intrinsically motivated model learning for developing curious robots", *Artificial Intelligence*, 2015.
[14] K. Subramanian, C. L. Isbell Jr, and A. L. Thomaz, "Exploration from demonstration for interactive reinforcement learning", in *AAMAS*, 2016.
[15] Y. Duan, X. Chen, R. Houthooft, J. Schulman, and P. Abbeel, "Benchmarking deep reinforcement learning for continous control", in *ICML*, 2016.
[16] J. Schmidhuber, "Curious model-building control systems", in *IJCNN*, 1991, pp. 1458–1463.
[17] Y. Sun, F. Gomez, and J. Schmidhuber, "Planning to be surprised: Optimal Bayesian exploration in dynamic environments", in *Artificial General Intelligence*, 2011, pp. 41–51.
[18] L. Itti and P. F. Baldi, "Bayesian surprise attracts human attention", in *NIPS*, 2005, pp. 547–554.
[19] A. Graves, "Practical variational inference for neural networks", in *NIPS*, 2011, pp. 2348–2356.
[20] C. Blundell, J. Cornebise, K. Kavukcuoglu, and D. Wierstra, "Weight uncertainty in neural networks", in *ICML*, 2015.
[21] J. Schmidhuber, "Formal theory of creativity, fun, and intrinsic motivation (1990–2010)", *IEEE Trans. Auton. Mental Develop.*, vol. 2, no. 3, pp. 230–247, 2010.
[22] J. Storck, S. Hochreiter, and J. Schmidhuber, "Reinforcement driven information acquisition in non-deterministic environments", in *ICANN*, vol. 2, 1995, pp. 159–164.
[23] J. Z. Kolter and A. Y. Ng, "Near-Bayesian exploration in polynomial time", in *ICML*, 2009, pp. 513–520.
[24] G. E. Hinton and D. Van Camp, "Keeping the neural networks simple by minimizing the description length of the weights", in *COLT*, 1993, pp. 5–13.
[25] J. Schmidhuber, "Simple algorithmic principles of discovery, subjective beauty, selective attention, curiosity & creativity", in *Intl. Conf. on Discovery Science*, 2007, pp. 26–38.
[26] D. P. Kingma, T. Salimans, and M. Welling, "Variational dropout and the local reparameterization trick", in *NIPS*, 2015, pp. 2575–2583.
[27] R. J. Williams, "Simple statistical gradient-following algorithms for connectionist reinforcement learning", *Mach. Learn.*, vol. 8, no. 3-4, pp. 229–256, 1992.
[28] J. Kober and J. R. Peters, "Policy search for motor primitives in robotics", in *NIPS*, 2009, pp. 849–856.
[29] J. Peters and S. Schaal, "Reinforcement learning by reward-weighted regression for operational space control", in *ICML*, 2007, pp. 745–750.
[30] P. Auer, T. Jaksch, and R. Ortner, "Near-optimal regret bounds for reinforcement learning", in *NIPS*, 2009, pp. 89–96.
[31] A. Guez, N. Heess, D. Silver, and P. Dayan, "Bayes-adaptive simulation-based search with value function approximation", in *NIPS*, 2014, pp. 451–459.
[32] R. S. Sutton, *Introduction to reinforcement learning*.
[33] S. Still and D. Precup, "An information-theoretic approach to curiosity-driven reinforcement learning", *Theory Biosci.*, vol. 131, no. 3, pp. 139–148, 2012.
[34] D. Y. Little and F. T. Sommer, "Learning and exploration in action-perception loops", *Closing the Loop Around Neural Systems*, p. 295, 2014.
[35] S. B. Thrun, "Efficient exploration in reinforcement learning", Tech. Rep., 1992.
[36] M. Lopes, T. Lang, M. Toussaint, and P.-Y. Oudeyer, "Exploration in model-based reinforcement learning by empirically estimating learning progress", in *NIPS*, 2012, pp. 206–214.
[37] J. Schossau, C. Adami, and A. Hintze, "Information-theoretic neuro-correlates boost evolution of cognitive systems", *Entropy*, vol. 18, no. 1, p. 6, 2015.
[38] K. Zahedi, G. Martius, and N. Ay, "Linear combination of one-step predictive information with an external reward in an episodic policy gradient setting: A critical analysis", *Front. Psychol.*, vol. 4, 2013.
[39] P.-Y. Oudeyer and F. Kaplan, "What is intrinsic motivation? a typology of computational approaches", *Front Neurorobot.*, vol. 1, p. 6, 2007.
[40] G. Montufar, K. Ghazi-Zahedi, and N. Ay, "Information theoretically aided reinforcement learning for embodied agents", *ArXiv preprint arXiv:1605.09735*, 2016.
[41] S. Mohamed and D. J. Rezende, "Variational information maximisation for intrinsically motivated reinforcement learning", in *NIPS*, 2015, pp. 2116–2124.
[42] C. Salge, C. Glackin, and D. Polani, "Guided self-organization: Inception", in 2014, ch. Empowerment–An Introduction, pp. 67–114.

## A  Bayesian neural networks (BNNs)

We demonstrate the behavior of a BNN [1] when trained on simple regression data. Figure 1 shows a snapshot of the behavior of the BNN during training. In this figure, the red dots represent the regression training data, which has a 1-dim input $x$ and a 1-dim output. The input to the BNN is constructed as $x = [x, x^2, x^3, x^4]$. The green dots represent BNN predictions, each for a differently sampled $\theta$ value, according to $q(\cdot; \phi)$. The color lines represent the output for different, but fixed, $\theta$ samples. The shaded areas represent the sampled output mean plus-minus one and two standard deviations.

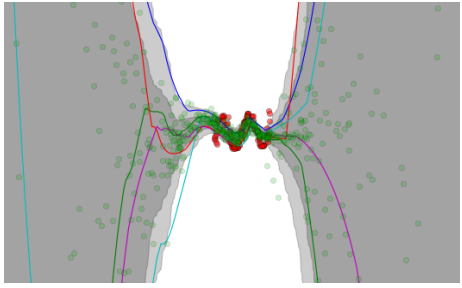
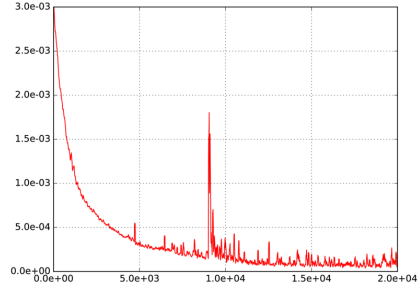

Figure 1: BNN output on a 1D regression task. Shaded areas: sampled output mean $\pm$ one/two standard deviations. Red dots: targets; green dots: prediction samples. Colored lines: neural network functions for different $\theta \sim q(\cdot; \phi)$ samples.

Figure 2: Just before iteration 10,000 we introduce data outside the training data range to the BNN. This results in a KL divergence spike, showing the model's surprise.

The figure shows that the BNN output is very certain in the training data range, while having high uncertainty otherwise. If we introduce data outside of this training range, or data that is significantly different from the training data, it will have a high impact on the parameter distribution $q(\theta; \phi)$. This is tested in Figure 2: previously unseen data is introduced right before training iteration 10,000. The KL divergence from posterior to prior (y-axis) is set out in function of the training iteration number (x-axis). We see a sharp spike in the KL divergence curve, which represents the BNN's surprise about this novel data. This spike diminishes over time as the BNN learns to fit this new data, becoming less surprised about it.

## B  Experimental setup

In case of the classic tasks CartPole, CartPoleSwingup, DoublePendulum, and MountainCar, as well as in the case of the hierarchical task SwimmerGather, the dynamics BNN has one hidden layer of 32 units. For the locomotion tasks Walker2D and HalfCheetah, the dynamics BNN has two hidden layers of 64 units each. All hidden layers have rectified linear unit (ReLU) nonlinearities, while no nonlinearity is applied to the output layer. The number of samples drawn to approximate the variational lower bound expectation term is fixed to 10. The batch size for the policy gradient methods is set to 5,000 samples, except for the SwimmerGather task, where it is set to 50,000. The replay pool has a fixed size of 100,000 samples, with a minimum size of 500 samples for all but the SwimmerGather task. In this latter case, the replay pool has a size of 1,000,000 samples. The dynamics BNN is updated each epoch, using 500 iterations of Adam [2], with a batch size of 10, except for the SwimmerGather task, in which 5,000 iterations are used. The Adam learning rate is set to 0.0001 while the batches are drawn randomly with replacement from the replay pool. In the second-order KL divergence update step, $\lambda$ is set to 0.01. The BNN prior weight distribution is a fully factorized Gaussian with $\mu$ sampled from a different Gaussian distribution $\mathcal{N}(\mathbf{0}, I)$, while $\rho$ is fixed to $\log(1 + e^{0.5})$.

The classic tasks make use of a neural network policy with one layer of 32 tanh units, while the locomotion tasks make use of a two-layer neural network of 64 and 32 tanh units. The outputs are modeled by a fully factorized Gaussian distribution $\mathcal{N}(\mu, \sigma^2 I)$, in which $\mu$ is modeled as the network output, while $\sigma$ is a parameter. The classic tasks make use of a neural network baseline with one layer of 32 ReLU units, while the locomotion tasks make use linear baseline function.

All tasks are implemented as described in [3]. The tasks have the following state and action dimensions: CartPole, $\mathcal{S} \subseteq \mathbb{R}^4, \mathcal{A} \subseteq \mathbb{R}^1$; CartPoleSwingup, $\mathcal{S} \subseteq \mathbb{R}^4, \mathcal{A} \subseteq \mathbb{R}^1$; DoublePendulum, $\mathcal{S} \subseteq \mathbb{R}^6, \mathcal{A} \subseteq \mathbb{R}^1$; MountainCar $\mathcal{S} \subseteq \mathbb{R}^3, \mathcal{A} \subseteq \mathbb{R}^1$; locomotion tasks HalfCheetah, $\mathcal{S} \subseteq \mathbb{R}^{20}, \mathcal{A} \subseteq \mathbb{R}^6$; and Walker2D, $\mathcal{S} \subseteq \mathbb{R}^{20}, \mathcal{A} \subseteq \mathbb{R}^6$; and hierarchical task SwimmerGather, $\mathcal{S} \subseteq \mathbb{R}^{33}, \mathcal{A} \subseteq \mathbb{R}^2$. The time horizon is set to $T = 500$ for all tasks. For the sparse reward experiments, the tasks have been modified as follows. In MountainCar, the agent receives a reward of $+1$ when the goal state is reached, namely escaping the valley from the right side. In CartPoleSwingup, the agent receives a reward of $+1$ when $\cos(\beta) > 0.8$, with $\beta$ the pole angle. Therefore, the agent has to figure out how to swing up the pole in the absence of any initial external rewards. In HalfCheetah, the agent receives a reward of $+1$ when $x_{\text{body}} > 5$. As such, it has to figure out how to move forward without any initial external reward.

# References

[1] C. Blundell, J. Cornebise, K. Kavukcuoglu, and D. Wierstra, "Weight uncertainty in neural networks", in *ICML*, 2015.

[2] D. Kingma and J. Ba, "Adam: A method for stochastic optimization", in *ICLR*, 2015.

[3] Y. Duan, X. Chen, R. Houthooft, J. Schulman, and P. Abbeel, "Benchmarking deep reinforcement learning for continous control", in *ICML*, 2016.